\documentclass[11pt,a4paper]{article}
\usepackage[hyperref]{acl2020}
\usepackage{times}
\usepackage{latexsym}
\usepackage{graphicx}
\usepackage{xcolor}
\usepackage{multirow}

\usepackage{amsmath}
\usepackage{amssymb}

\renewcommand{\paragraph}[1]{\noindent{\bf{#1}}}

\usepackage{microtype}

\aclfinalcopy 
\def\aclpaperid{2223} 

\title{
\vspace*{-0.5in}
{{\small \hfill ACL'2020}\\
\vspace*{.25in}} 
Cross-Modality Relevance for Reasoning on Language and Vision}

\author{Chen Zheng, Quan Guo, Parisa Kordjamshidi \\
        Department of Computer Science and Engineering, Michigan State University \\
        East Lansing, MI 48824 \\
        \texttt{\{zhengc12,guoquan,kordjams\}@msu.edu} \\
}

\date{}

\begin{document}
\maketitle
\begin{abstract}

This work deals with the challenge of learning and reasoning over language and vision data for the related downstream tasks such as \textit{visual question answering (VQA)} and \textit{natural language for visual reasoning (NLVR)}.
We design a novel cross-modality relevance module that is used in an end-to-end framework to learn the relevance representation between components of various input modalities under the supervision of a target task, which is more generalizable to unobserved data compared to merely reshaping the original representation space.
In addition to modeling the relevance between the textual entities and visual entities, we model the higher-order relevance between entity relations in the text and object relations in the image. 
Our proposed approach shows competitive performance on two different language and vision tasks using public benchmarks and improves the state-of-the-art published results. The learned alignments of input spaces and their relevance representations by NLVR task boost the training efficiency of VQA task.

\end{abstract}

\section{Introduction}

Real-world problems often involve data from multiple modalities and resources. Solving a problem at hand usually requires the ability to reason about the components across all the involved modalities. Examples of such tasks are visual question answering~(VQA)~\cite{antol2015vqa,vqa2} and natural language visual reasoning~(NLVR)~\cite{suhr-etal-2017-corpus,Suhr2018ACF}.
One key to intelligence here is to identify the relations between the modalities, combine and reason over them for decision making.
Deep learning is a prominent technique to learn representations of the data for decision making for various target tasks. It has achieved supreme performance based on large scale corpora~\cite{devlin-etal-2019-bert}.
However, it is a challenge to learn joint representations for cross-modality data because deep learning is data-hungry.
There are many recent efforts to build such multi-modality datasets~\cite{lin2014microsoft,krishna2017visual,Johnson_2017_CVPR,antol2015vqa,suhr-etal-2017-corpus,vqa2,Suhr2018ACF}.
Researchers develop models by joining features, aligning representation spaces, and using Transformers~\cite{li2019visualbert,tan2019lxmert}. However, generalizability is still an issue when operating on unobserved data. It is hard for deep learning models to capture high-order patterns of reasoning, which is essential for their generalizability.

There are several challenging research directions for addressing learning representations for cross-modality data and enabling reasoning for target tasks. First is the alignment of the representation spaces for multiple modalities; second is designing architectures with the ability to capture high-order relations for generalizability of reasoning; third is using pre-trained modules to make the most use of minimal data.

An orthogonal direction to the above-mentioned aspects of learning is finding relevance between the components and the structure of various modalities when working with multi-modal data.
Most of the previous language and visual reasoning models try to capture the relevance by learning representations based on an attention mechanism. 
Finding relevance, known as matching, is a fundamental task in information retrieval~(IR)~\cite{mitra2017learning}. Benefiting from matching, Transformer models gain the excellent ability to index, retrieve, and combine features of underlying instances by a matching score~\cite{vaswani2017attention}, which leads to the state-of-the-art performance in various tasks~\cite{devlin-etal-2019-bert}. However, the matching in the attention mechanism is used to learn a set of weights to highlight the importance of various components.

In our proposed model, we learn representations directly based on the relevance score inspired by the ideas from IR models. In contrast to the attention mechanism and Transformer models, we claim that {\bf the relevance patterns are as important}. With proper alignment of the representation spaces of different input modalities, matching can be applied to those spaces. 
The idea of learning relevance patterns is similar to Siamese networks~\cite{Koch2015SiameseNN} which learn transferable patterns of similarity of two image representations for one-shot image recognition.
Similarity metric between two modalities is shown to be helpful for aligning multiple spaces of modalities~\cite{Frome2013DeViSEAD}.

The contributions of this work are as follows: {\bf 1)} We propose a cross-modality relevance~(CMR) framework that considers entity relevance and high-order relational relevance between the two modalities with an alignment of representation spaces. The model can be trained end-to-end with customizable target tasks. {\bf 2)} We evaluate the methods and analyze the results on both VQA and NLVR tasks using VQA v2.0 and $\mbox{NLVR}^2$ datasets respectively. We improve state-of-the-art on both tasks' published results. Our analysis shows the significance of the patterns of relevance for the reasoning, and the CMR model trained on $\mbox{NLVR}^2$ boosts the training efficiency of the VQA task.

\section{Related Work}

\paragraph{Language and Vision Tasks.}
Learning and decision making based on natural language and visual information has attracted the attention of many researchers due to exposing many interesting research challenges to the AI community. 
Among many other efforts~\cite{lin2014microsoft,krishna2017visual,Johnson_2017_CVPR}, \citeauthor{antol2015vqa} proposed the VQA challenge that contains open-ended questions about images that require an understanding of and reasoning about language and visual components. \citeauthor{Suhr2018ACF} proposed the NLVR task that asks models to determine whether a sentence is true based on the image. 

\paragraph{Attention Based Representation.}
Transformers are stacked self-attention models for general purpose sequence representation~\cite{vaswani2017attention}. They have been shown to achieve extraordinary success in natural language processing not only for better results but also for efficiency due to their parallel computations. Self-attention is a mechanism to reshape representations of components based on relevance scores. They have been shown to be effective in generating contextualized representations for text entities. More importantly, there are several efforts to pre-train huge Transformers based on large scale corpora~\cite{devlin-etal-2019-bert,Yang2019XLNetGA,radford2019language} on multiple popular tasks to enable exploiting them and performing other tasks with small corpora.
Researchers also extended Transformers with both textual and visual modalities~\cite{li2019visualbert,Sun2019VideoBERTAJ,tan2019lxmert,Su2020VL-BERT,Tsai2019MultimodalTF}. Sophisticated pre-training strategies were introduced to boost the performance~\cite{tan2019lxmert}.
However, as mentioned above, modeling relations between components is still a challenge for the approaches that try reshaping the entity representation space while the relevance score can be more expressive for these relations. In our CMR framework, we model high-order relations in relevance representation space rather than the entity representation space.

\paragraph{Matching Models.}
Matching is a fundamental task in information retrieval~(IR). There are IR models that focus on comparing the global representation matching~\cite{huang2013learning,shen2014learning}, the local components ({\it a.k.a} terms) matching~\cite{guo2016deep,pang2016text}, and hybrid methods~\cite{mitra2017learning}.
Our relevance framework is partially inspired by the local components matching which we apply here to model the relevance of the components of the model's inputs.
However, our work differs in several significant ways. First, we work under the cross-modality setting. Second, we extend the relevance to a high-order, {\it i.e.} model the relevance of entity relations. Third, our framework can work with different target tasks, and we show that the parameters trained on one task can boost the training of another.

\begin{figure*}
\centering
\includegraphics[width=1.0\textwidth]{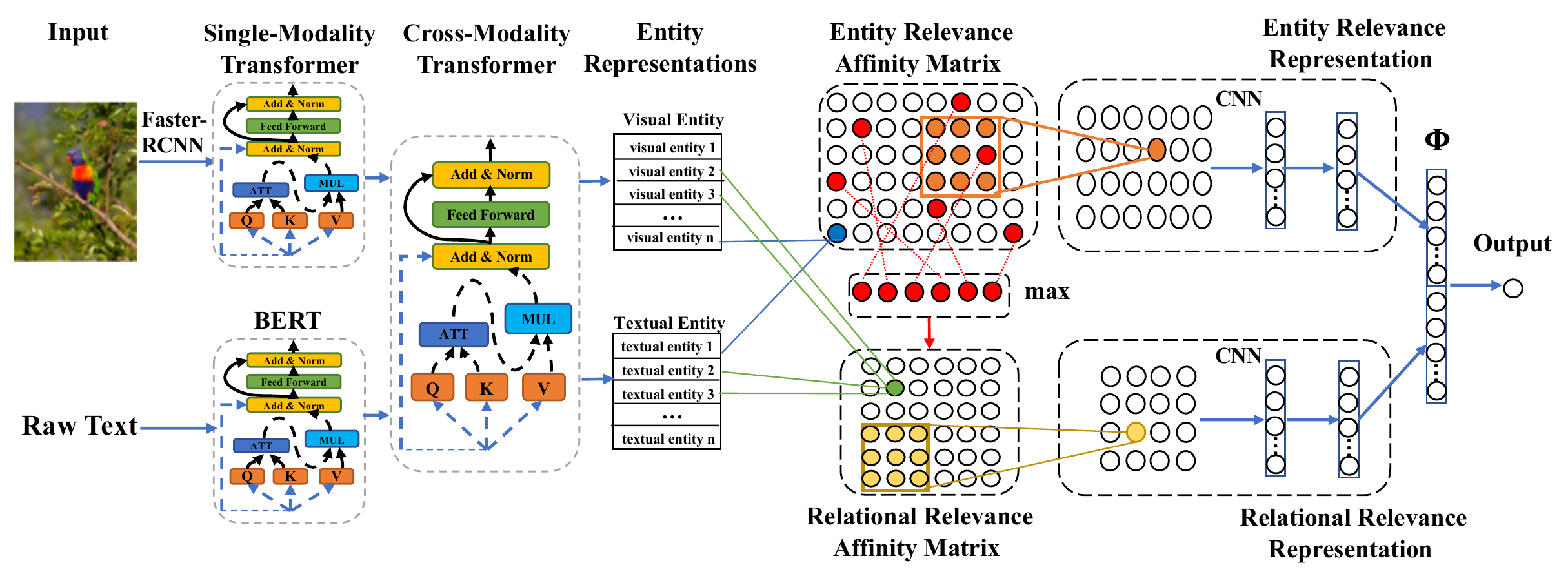}
\caption{Cross-Modality Relevance model is composed of single-modality transformer, cross-modality transformer,  entity relevance, and high-order relational relevance, followed by a task-specific classifier.\label{fig:architecture}}
\end{figure*}

\section{Cross-Modality Relevance}

Cross-Modality Relevance~(CMR) aims to establish a framework for general purpose relevance in various tasks. As an end-to-end model, it encodes the relevance between the components of input modalities under task-specific supervision. We further add a high-order relevance between relations that occur in each modality.

Figure~\ref{fig:architecture} shows the proposed architecture. We first encode data from different modalities with single modality Transformers and align the encoding spaces by a cross-modality Transformer. We consistently refer to the words in text and objects in images  ({\it i.e.} bounding boxes in images) as ``entities'' and their representations as ``Entity Representations''. We use the relevance between the components of the two modalities to model the relation between them. The relevance includes the relevance between their entities, as shown in the ``Entity Relevance'', and high-order relevance between their relations, as shown in the ``Relational Relevance''. We learn the representations of the affinity matrix of relevance score by convolutional layers and fully-connected layers. Finally, we predict the output by a non-linear mapping based on all the relevance representations.
This architecture can help to solve tasks that need reasoning on two modalities based on their relevance. We argue that the parameters trained on one task can boost the training of the other tasks that deal with multi-modality reasoning.

In this section, we first formulate the problem. Then we describe our cross-modality relevance~(CMR) model for solving the problem. The architecture, loss function, and training procedure of CMR are explained in detail. We will use the VQA and NLVR tasks as showcases.

\subsection{Problem Formulation}

Formally, the problem is to model a mapping from a cross-modality data sample $\mathcal{D}=\left\{\mathcal{D}_\mu\right\}$ to an output $y$ in a target task, where $\mu$ denotes the type of modality. And $\mathcal{D}_\mu=\left\{d^\mu_1, \cdots, d^\mu_{N^\mu}\right\}$ is a set of entities in the modality $\mu$.
In visual question answering, VQA, the task is to predict an answer given two modalities, that is a textual question ($\mathcal{D}_t$) and a visual image ($\mathcal{D}_v$). In NLVR,  given a textual statement ($\mathcal{D}_t$) and an image ($\mathcal{D}_v$), the task is to determine the correctness of the textual statement.

\subsection{Representation Spaces Alignment}

\paragraph{Single Modality Representations.}
For the textual modality $\mathcal{D}_t$, we utilize BERT~\cite{devlin-etal-2019-bert} as shown in the bottom-left part of Figure~\ref{fig:architecture},
which is a multi-layer Transformer~\cite{vaswani2017attention} with three different inputs: WordPieces embeddings~\cite{wu2016google}, segment embeddings, and position embeddings. We refer to all the words as the entities in the textual modality and use the BERT representations for textual single-modality representations $\left\{s^t_1,\cdots, s^t_{N^t}\right\}$. We assume to have $N^t$ words as textual entities.

For visual modality $\mathcal{D}_v$,
as shown in the top-left part of Figure~\ref{fig:architecture},
Faster-RCNN~\cite{ren2015faster} is used to generate regions of interest~(ROIs), extract dense encoding representations of the ROIs, and predict the probability of each ROI.
We refer to the ROIs on images as the visual entities $\left\{d^v_1,\cdots,d^v_{N^v}\right\}$. We consider a fixed number, $N^v$, of visual entities with highest probabilities predicted by Faster-RCNN each time.
The dense representation of each ROI is a local latent representation of a $2048$-dimensional vector~\cite{ren2015faster}. To enrich the visual entity representation with the visual context, we further project the vectors with feed-forward layers and encode them by a single-modality Transformer as shown in the second column in Figure~\ref{fig:architecture}. The visual Transformer takes the dense representation, segment embedding, and pixel position embedding~\cite{tan2019lxmert} as input and generates the single-modality representation $\left\{s^v_1,\cdots,s^v_{N^v}\right\}$.
In case there are multiple images, for example, NLVR data ($\mbox{NLVR}^2$) has two images in each example, each image is encoded by the same procedure and we keep $N^v$ visual entities per image. We refer to this as different sources of the same modality throughout the paper. We restrict all the single-modality representations to be vectors of the same dimension $d$.
However, these original representation spaces should be aligned.

\paragraph{Cross-Modality Alignment.}
To align the single-modality representations in a uniformed representation space, we introduce a cross-modality Transformer as shown in the third column of Figure~\ref{fig:architecture}. All the entities are treated uniformly in the modality Transformer. Given the set of entity representations from all modalities we define the matrix with all the elements in the set $S=\left[s^t_1, \cdots, s^t_{N^t}, s^v_1, \cdots, s^v_{N^v}\right]\in\mathbf{R}^{d\times (N^t+N^v)}$. Each cross-modality self-attention calculation is computed as follows~\cite{vaswani2017attention}\footnote{Please note here we keep the usual notation of the attention mechanism for this equation. The notations might have been overloaded in other parts of the paper.}, 
\begin{equation}\mbox{Attention}\left(K,Q,V\right)
    = \mbox{softmax}\left(\frac{K^{\top} Q}{\sqrt{d}}\right) V,
\end{equation}
where in our case the key $K$, query $Q$, and value $V$, all are the same tensor $S$, and $\mbox{softmax}\left(\cdot\right)$ normalizes along the columns.
A cross-modality Transformer layer consists of a cross-modality self-attention representation followed by residual connection with normalization from the input representation, a feed-forward layer, and another residual connection normalization. We stack several cross-modality Transformer layers to get a uniform representation over all modalities. We refer to the resulting uniformed representations as the entity representation and denote the set of the entity representations of all the entities as $\left\{s^{'t}_1, \cdots, s^{'v}_{N^t}, s^{'v}_1, \cdots, s^{'v}_{N^v}\right\}$. Although the representations are still organized by their original modalities per entity, they carry the information from the interactions with the other modality and are aligned in uniform representation space. The entity representations, as the fourth column in Figure~\ref{fig:architecture}, alleviate the gap between representations from different modalities, as we will show in the ablation studies, and allow them to be matched in the following steps.

\subsection{Entity Relevance}
\label{entity_rel}

Relevance plays a critical role in reasoning ability, which is required in many tasks such as information retrieval, question answering, intra- and inter-modality reasoning.
Relevance patterns are independent from input representation space, and can have better generalizability to unobserved data.
To consider the entity relevance between two modalities $\mathcal{D}_\mu$ and $\mathcal{D}_\nu$, the entity relevance representation is calculated as shown in Figure~\ref{fig:architecture}.
Given entity representation matrices $S^{'\mu}=\left[s^{'\mu}_1, \cdots, s^{'\mu}_{N^\mu}\right]\in\mathbf{R}^{d\times N^\mu}$ and $S^{'\nu}=\left[s^{'\nu}_1, \cdots, s^{'\nu}_{N^\nu}\right]\in\mathbf{R}^{d\times N^\nu}$, the relevance representation is calculated by
\begin{subequations}
\begin{align}
    A^{\mu,\nu} &= \left(S^{'\mu}\right)^{\top} S^{'\nu}, \label{eq:affinity} \\
    \mbox{M}\left( \mathcal{D}_\mu,\mathcal{D}_\nu \right) &= \mbox{CNN}_{\mathcal{D}_\mu,\mathcal{D}_\nu}\left(A^{\mu,\nu} \right),
\end{align}
\end{subequations}
where $A^{\mu,\nu}$ is the affinity matrix of the two modalities as shown in the right side of Figure~\ref{fig:architecture}. $A^{\mu,\nu}_{ij}$ is the relevance score of $i$th entity in $\mathcal{D}_\mu$ and $j$th entity in $\mathcal{D}_\nu$.
$\mbox{CNN}_{\mu,\nu}\left(\cdot\right)$ is a CNN, corresponding to the sixth column of Figure~\ref{fig:architecture}, which contains several convolutional layers and fully connected layers. Each convolutional layer is followed by a max-pooling layer. Fully connected layers finally map the flatten feature maps to $d$-dimensional vector.
We refer to $\Phi_{\mathcal{D}_\mu,\mathcal{D}_\nu}=\mbox{M}\left(\mathcal{D}_\mu,\mathcal{D}_\nu\right)$ as the entity relevance representation between $\mu$ and $\nu$.

We compute the relevance between different modalities. For the modalities considered in this work, when there are multiple images in the visual modality, we calculate the relevance representation between them too.
In particular, for VQA dataset, the above setting results in one entity relevance representation: a textual-visual entity relevance $\Phi_{\mathcal{D}_t,\mathcal{D}_v}$. For $\mbox{NLVR}^2$ dataset, there are three entity relevance representations: two textual-visual entity relevance $\Phi_{\mathcal{D}_t,\mathcal{D}_{v_1}}$ and $\Phi_{\mathcal{D}_t,\mathcal{D}_{v_2}}$, and a visual-visual entity relevance $\Phi_{\mathcal{D}_{v_1},\mathcal{D}_{v_2}}$ between two images.
Entity relevance representations will be flattened and joined with other features in the next layer of the network.

\subsection{Relational Relevance}

We also consider the relevance beyond entities, that is, the relevance of the entities' relations. This extension allows our CMR to capture higher-order relevance patterns.
We consider pair-wise non-directional relations between entities in each modality and calculate the relevance of the relations across modalities.
The procedure is similar to entity relevance as shown in Figure~\ref{fig:architecture}.
We denote the relational representation as a non-linear mapping $\mathbf{R}^{2d} \rightarrow \mathbf{R}^{d}$ modeled by fully-connected layers from the concatenation of representations of the entities in the relation $r^{\mu}_{(i,j)}=\mbox{MLP}_{\mu,1}\left(\left[s^{'\mu}_i,s^{'\mu}_j\right]\right) \in \mathbf{R}^d$.
Relational relevance affinity matrix can be calculated by matching the relational representation, $\left\{r^{\mu}_{(i,j)}, \forall i \neq j \right\}$, from different modalities.
However, there will be $C^2_{N_\mu}$ possible pairs in each modality $\mathcal{D}_\mu$, most of which are irrelevant. The relational relevance representations will be sparse because of the irrelevant pairs on both sides. Computing the relevance score of all possible pairs will introduce a large number of unnecessary parameters which makes the training more difficult.

\begin{figure}
\centering
\includegraphics[width=1.0\columnwidth]{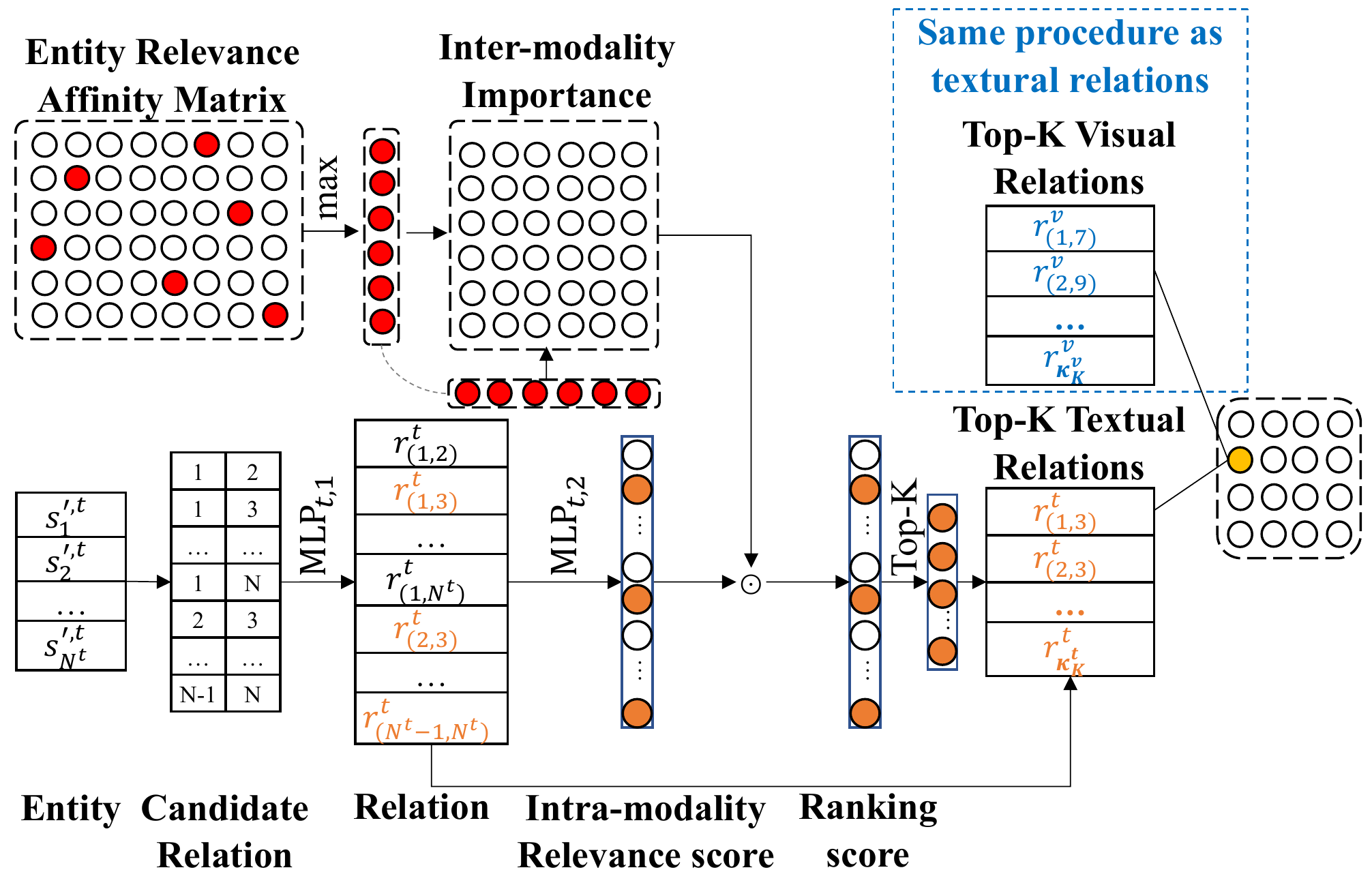}
\caption{Relational Relevance is the relevance of top-K relations in terms of intra-modality relevance score and inter-modality importance.}\label{fig:rel_match}
\end{figure}

We propose to rank the relation candidates (i.e. pairs) by the intra-modality relevance score and the inter-modality importance. Then we compare the top-$K$ ranked relation candidates between two modalities as shown in Figure~\ref{fig:rel_match}.
For the intra-modality relevance score, shown in the bottom left part of the figure, we estimate a normalized score based on the relational representation by a softmax layer.
\begin{equation}
    U^{\mu}_{(i,j)} = \frac{\exp \left( \mbox{MLP}_{\mu,2}\left(r^{\mu}_{(i,j)}\right)\right)}
    {\sum_{k\neq l} \exp\left(\mbox{MLP}_{\mu,2}\left(r^{\mu}_{(k,l)}\right)\right)}.
\end{equation}

To evaluate the inter-modality importance of a relation candidate, which is a pair of entities in the same modality, we first compute the relevance of each entity in text with respect to the visual objects. As shown in Figure~\ref{fig:rel_match}, we project a vector that includes the most relevant visual object for each word, denoted this importance vector as $v^{t}$. This helps to focus on words that are grounded in the visual modality. We use the same procedure to compute the most relevant words to each visual object. 

Then we calculate the relation candidates importance matrix $V^{\mu}$ by an outer product, $\otimes$, of the importance vectors as follows, 
\begin{subequations}
\begin{align}
    v^{\mu}_i &= \max_{j} A^{\mu, \nu}_{ij}, \\
    V^{\mu} &= v^{\mu} \otimes v^{\mu},
\end{align}
\end{subequations}
where $v^{\mu}_i$ is the $i$th scalar element in $v^{\mu}$ that corresponds to the $i$th entity, and $A^{\mu,\nu}$ is the affinity matrix calculated by Equation~\ref{eq:affinity}.

Notice that the inter-modality importance $V^{\mu}$ is symmetric. The upper triangular part of $V^{\mu}$, excluding the diagonal, indicates the importance of the corresponding elements with the same index in intra-modality relevance scores $U^{\mu}$. The ranking score for the candidates is the combination (here the product) of the two scores $W^{\mu}_{(i,j)} = U^{\mu}_{(i,j)} \times V^{\mu}_{ij}$. We select the set of top-$K$ ranked candidate relations $\mathcal{K}_\mu=\left\{\kappa_1, \kappa_2, \cdots, \kappa_K\right\}$. We reorganize the representation of the top-$K$ relations as $R^{\mu}=\left[r^{\mu}_{\kappa_1}, \cdots r^{\mu}_{\kappa_K}\right] \in \mathbf{R}^{d\times K}$. The relational relevance representation between $\mathcal{K}_\mu$ and $\mathcal{K}_\nu$ can be calculated similar to the entity relevance representations as shown in Figure~\ref{fig:architecture}.
\begin{equation}
    \mbox{M}\left(\mathcal{K}_\mu,\mathcal{K}_\nu\right) = \mbox{CNN}_{\mathcal{K}_\mu,\mathcal{K}_\nu}\left( \left(R^{\mu}\right)^{\top} R^{\nu} \right).
\end{equation}
$\mbox{M}\left(\mathcal{K}_\mu,\mathcal{K}_\nu\right)$ has its own parameters which results in a $d$-dimensional feature space $\Phi_{\mathcal{K}_\mu,\mathcal{K}_\nu}$. 

In particular, for VQA task, the above setting results in one relational relevance representation: a textual-visual relevance $\mbox{M}\left(\mathcal{K}_t,\mathcal{K}_v\right)$. For NLVR task, there are three entity relevance representations: two textual-visual relational relevance $\mbox{M}\left(\mathcal{K}_t,\mathcal{K}_{v_1}\right)$ and $\mbox{M}\left(\mathcal{K}_t,\mathcal{K}_{v_2}\right)$, and a visual-visual relational relevance $\mbox{M}\left(\mathcal{K}_{v_1},\mathcal{K}_{v_2}\right)$ between two images.
Relational relevance representations will be flattened and joined with other features in the next layers of the network.

After acquiring all the entity and relational relevance representations, namely $\Phi_{\mathcal{D}_\mu,\mathcal{D}_\nu}$ and $\Phi_{\mathcal{K}_\mu,\mathcal{K}_\nu}$, we concatenate them and use the result as the final feature $\Phi=\left[ \Phi_{\mathcal{D}_\mu,\mathcal{D}_\nu}, \cdots, \Phi_{\mathcal{K}_\mu,\mathcal{K}_\nu}, \cdots\right]$. A task-specific classifier $\mbox{MLP}_\Phi\left(\Phi\right)$ predicts the output of the target task as shown in the right-most column in Figure~\ref{fig:architecture}.

\subsection{Training}
{\bf End-to-end Training.}
CMR can be considered as an end-to-end relevance representation extractor. We simply predict the output $y$ from a specific task with the final feature $\Phi$ with a differentiable regression or classification function. 
The gradient of the loss function is back-propagated to all the components in CMR to penalize the prediction and adjust the parameters.
We freeze the parameters of the basic feature extractors, namely BERT for textual modality and Faster-RCNN for visual modality. The parameters of the following parts will be updated by gradient descent: single modality Transformers (except BERT), the cross-modality Transformers, $\mbox{CNN}_{\mathcal{D}_\mu,\mathcal{D}_\nu}\left(\cdot\right)$, $\mbox{CNN}_{\mathcal{K}_\mu,\mathcal{K}_\nu}\left(\cdot\right)$, $\mbox{MLP}_{\mu,1}\left(\cdot\right)$,  $\mbox{MLP}_{\mu,2}\left(\cdot\right)$ for all modalities and modality pairs, and the task-specific classifier $\mbox{MLP}_\Phi\left(\Phi\right)$.

The VQA task can be formulated as a multi-class classification that chooses a word to answer the question. We apply a softmax classifier on $\Phi$ and penalize with the cross-entropy loss.
For $\mbox{NLVR}^2$ dataset, the task is binary classification that determines whether the statement is correct regarding the images. We apply a logistic regression on $\Phi$ and penalize with the cross-entropy loss.

\paragraph{Pre-training Strategy.}
To leverage the pre-trained parameters of our cross-modality Transformer and relevance representations, we use the following training settings.
For all tasks, we freeze the parameters in BERT and faster-RCNN. We used pre-trained parameters in the (visual) single modality Transformers as proposed by~\cite{tan2019lxmert} and leave them being fine-tuned with the following procedure. Then we randomly initialize and train all the parameters in the model on NLVR with $\mbox{NLVR}^2$ dataset. After that, we keep and fine-tune all the parameters on the VQA task with the VQA v2.0 dataset. (See data description Section~\ref{subset:data}.) 
In this way, the parameters of the cross-modality Transformer and relevance representations, pre-trained by $\mbox{NLVR}^2$ dataset, are reused and fine-tuned on the VQA dataset. Only the final task-specific classifier with the input features $\Phi$ is initialized randomly. The pre-trained cross-modality Transformer and relevance representations help the model for VQA to converge faster and achieve a competitive performance compared to the state-of-the-art results.

\section{Experiments and Results}


\subsection{Data Description}\label{subset:data}

\paragraph{$\mbox{NLVR}^2$}~\cite{Suhr2018ACF} is a dataset that aims to joint reasoning about natural language descriptions and related images. Given a textual statement and a pair of images, the task is to indicate whether the statement correctly describes the two images. $\mbox{NLVR}^2$ contains $107,292$ examples of sentences paired with visual images and designed to emphasize semantic diversity, compositionality, and visual reasoning challenges.

\paragraph{VQA v2.0}~\cite{vqa2} is an extended version of the VQA dataset. It contains $204,721$ images from the MS COCO~\cite{lin2014microsoft}, paired with $1,105,904$ free-form, open-ended natural language questions and answers. These questions are divided into four categories: Yes/No, Number, and Other.

\subsection{Implementation Details}\label{subsec:impl}
We implemented CMR using Pytorch\footnote{Our code and data is
available at \url{https://github.com/HLR/Cross_Modality_Relevance}.}. We consider the $768$-dimension single-modality representations. For textural modality, the pre-trained BERT ``base'' model~\cite{devlin-etal-2019-bert} is used to generate the single-modality representation. For visual modality, we use Faster-RCNN pre-trained by \citeauthor{anderson2018bottom}, followed by a five-layers Transformer. Parameters in BERT and Faster-RCNN are fixed.
For each example, we keep $20$ words as textual entities and $36$ ROIs per image as visual entities. For the relational relevance, top-10 ranked pairs are used. For each relevance CNN, $\mbox{CNN}_{\mathcal{D}_\mu,\mathcal{D}_\nu}\left(\cdot\right)$ and $\mbox{CNN}_{\mathcal{K}_\mu,\mathcal{K}_\nu}\left(\cdot\right)$, we use two convolutional layers, each of which is followed by a max-pooling, and fully connected layers.
For the relational representations and their intra-modality relevance score, $\mbox{MLP}_{\mu,1}\left(\cdot\right)$ and  $\mbox{MLP}_{\mu,2}\left(\cdot\right)$, we use one hidden layer for each. The task-specific classifier $\mbox{MLP}_\Phi\left(\Phi\right)$ contains three hidden layers.
The model is optimized using the Adam optimizer with $\alpha=10^{-4}, \beta_1=0.9, \beta_2=0.999, \epsilon=10^{-6}$. The model is trained with a weight decay $0.01$, a max gradient normalization $1.0$, and a batch size of 32.

\subsection{Baseline Description}



\paragraph{VisualBERT}~\cite{li2019visualbert} is an End-to-End model for language and vision tasks, consists of Transformer layers that align textual and visual representation spaces with self-attention. VisualBERT and CMR have a similar cross-modality alignment approach. However, VisualBERT only uses the Transformer representations while CMR uses the relevance representations.

\paragraph{LXMERT}~\cite{tan2019lxmert} aims to learn cross-modality encoder representations from Transformers. It pre-trains the model with a set of tasks and fine-tunes on another set of specific tasks. LXMERT is the currently published state-of-the-art on both $\mbox{NLVR}^2$ and VQA v$2.0$.

\subsection{Results}

\paragraph{$\mbox{NLVR}^2$:} The results of NLVR task are listed in Table~\ref{tab:nlvr_result}. Transformer based models~(VisualBERT, LXMERT, and CMR) outperform other models~(N2NMN~\cite{hu2017learning}, MAC~\cite{hudson2018compositional}, and FiLM~\cite{perez2018film}) by a large margin. This is due to the strong pre-trained single-modality representations and the Transformers' ability to reshape the representations that align the spaces.
Furthermore, CMR shows the best performance compared to all Transformer-based baseline methods and achieves state-of-the-art.  VisualBERT and CMR have similar cross-modality alignment approach. CMR outperforms VisualBERT by $12.4\%$. The gain mainly comes from entity relevance and relational relevance that model the relations.

\begin{table}
\begin{center}\small
\begin{tabular}{c|cc}
\hline
\textbf{Models}& Dev$\%$ & Test$\%$ \\
\hline
N2NMN            & 51.0 & 51.1   \\
MAC-Network      & 50.8	& 51.4   \\
FiLM             & 51.0 & 52.1   \\
CNN+RNN          & 53.4 & 52.4   \\
VisualBERT       & 67.4 & 67.0   \\
LXMERT           & 74.9 & 74.5   \\
\bf{CMR}        & \bf{75.4} & \bf{75.3}   \\
\hline
\end{tabular}
\end{center}
\caption{Accuracy on $\mbox{NLVR}^2$.}
\label{tab:nlvr_result}
\end{table}

\paragraph{VQA v2.0:}
In Table~\ref{tab:v2results}, we show the comparison with published models excluding the ensemble ones.
Most competitive models are based on Transformers (ViLBERT~\cite{lu2019vilbert}, VisualBERT~\cite{li2019visualbert}, VL-BERT~\cite{Su2020VL-BERT}, LXMERT~\cite{tan2019lxmert}, and CMR). BUTD~\cite{anderson2018bottom,teney2018tips}, ReGAT~\cite{Li2019RelationAwareGA}, and BAN~\cite{kim2018bilinear} also employ attention mechanism for a relation-aware model.
The proposed CMR achieves the best test accuracy on Y/N questions and Other questions.
However, CMR does not achieve the best performance on \textit{Number} questions. This is because Number questions require the ability to count numbers in one modality while CMR focuses on modeling relations between modalities. Performance on counting might be improved by explicit modeling of quantity representations. 
CMR also achieves the best overall accuracy. In particular, we can see a $2.3\%$ improvement over VisualBERT~\cite{li2019visualbert}, as in the above mentioned $\mbox{NLVR}^2$ results. This shows the significance of the entity and relational relevance.

\begin{table}
  \begin{center}\small
  \begin{tabular}{c|c|cccc}
  \hline
     \multirow{2}{*}{Model} & Dev$\%$ &\multicolumn{4}{c}{Test Standard$\%$} \\
     \cline{2-6} 
     & Overall & Y/N & Num & Other & Overall \\
  \hline
      BUTD & 65.32 &  81.82& 44.21& 56.05 & 65.67 \\
  \hline
      ReGAT & 70.27 &  86.08 & 54.42 & 60.33 & 70.58 \\
      ViLBERT & 70.55 & - & - & - & 70.92 \\
      VisualBERT & 70.80 & - & - & - & 71.00 \\
      BAN & 71.4 & 87.22 &	54.37 &	62.45 &	71.84 \\
      VL-BERT & 71.79 & 87.94 &	54.75 &	62.54 &	72.22 \\
      LXMERT & 72.5 & 87.97 &	\bf{54.94} &	63.13&	72.54 \\
  \hline
      \bf{CMR} & 72.58& \bf{88.14}&	54.71&	\bf{63.16}&	\bf{72.60} \\
  \hline
  \end{tabular}
  \end{center}
  \caption{Accuracy on VQA v2.0.}
  \label{tab:v2results}
\end{table}

Another observation is that, if we train CMR for VQA task from scratch with random initialization while still use the fixed BERT and Faster-RCNN, the model converges after 20 epochs. As we initialize the parameters with the model trained on $\mbox{NLVR}^2$, it takes 6 epochs to converge. The significant improvement of convergence speed indicates that the optimal model for VQA is close to that of NLVR.

\section{Analysis}
\subsection{Model Size}

To investigate the influence of model sizes, we empirically evaluated CMR on $\mbox{NLVR}^2$ with various sets of Transformers sizes which contain the most parameters of the model. All other details are kept the same as descriptions in Section~\ref{subsec:impl}. Textual Transformer remains 12 layers because it is the pre-trained BERT. Our model contains $285M$ parameters. Among these parameters, around $230M$ parameters belong to pre-trained BERT and Transformer. Table~\ref{tab:model_size} shows the results. As we increase the number of layers in the visual Transformer and the cross-modality Transformer, it tends to improve accuracy. However, the performance becomes stable when there are more than five layers. We choose five layers of visual Transformer and cross-modality Transformer in other experiments.

\begin{table}
\begin{center}\small
\begin{tabular}{ccc|cc}
\hline
Textural & Visual & Cross & Dev$\%$ & Test$\%$ \\
\hline
   12 & 3 & 3  & 74.1 & 74.4   \\
   12 & 4 & 4  & 74.9 & 74.7   \\
   12 & 5 & 5  & 75.4 & \textbf{75.3}   \\
   12 & 6 & 6  & 75.5 & 75.1   \\
\hline
\end{tabular}
\end{center}
\caption{Accuracy on $\mbox{NLVR}^2$ of CMR with various Transformer sizes. The numbers in the left part of the table indicate the number of self-attention layers.}\label{tab:model_size}
\end{table}

\subsection{Ablation Studies}

To better understand the influence of each part in CMR, we perform the ablation study. Table~\ref{tab:ablation} shows the performances of four variations on $\mbox{NLVR}^2$.

\begin{table}
\begin{center}\small
\begin{tabular}{c|cc}
\hline
\textbf{Models}& Dev$\%$ & Test$\%$ \\
\hline
\bf{CMR}        & \bf{75.4} & \bf{75.3}   \\
\hline
without Single-Modality Transformer     & 68.2 & 68.5     \\
without Cross-Modality Transformer  & 59.7 & 59.1  \\
without Entity Relevance           & 70.6 &  71.2   \\
without Relational Relevance           & 73.0 & 73.4   \\
\hline
\end{tabular}
\end{center}
\caption{Test accuracy of different variations of CMR on $\mbox{NLVR}^2$.}
\label{tab:ablation}
\end{table}

\paragraph{Effect of Single Modality Transformer.}
We remove both textual and visual single-modality Transformers and map the raw input with a linear transformation to $d$-dimensional space instead. Notice that the raw input of textual modality is the WordPieces~\cite{wu2016google} embeddings, segment embeddings, and the position embeddings of each word, while that of visual modality is the $2048$-dimension dense representation of each ROI extracted by Faster-RCNN. It turns out that removing single-modality Transformers decreases the accuracy by $9.0\%$. Single modality Transformers play a critical role in producing a strong contextualized representation for each modality.

\paragraph{Effect of Cross-Modality Transformer.}
We remove the cross-modality Transformer and use single-modality representations as entity representations. As shown in Table~\ref{tab:ablation}, the model degenerates dramatically, and the accuracy decreases by $16.2\%$. The huge accuracy gap demonstrates the unparalleled contribution of the cross-modality Transformer to aligning representation spaces from input modalities.

\begin{figure}
\centering
\includegraphics[width=1.0\columnwidth]{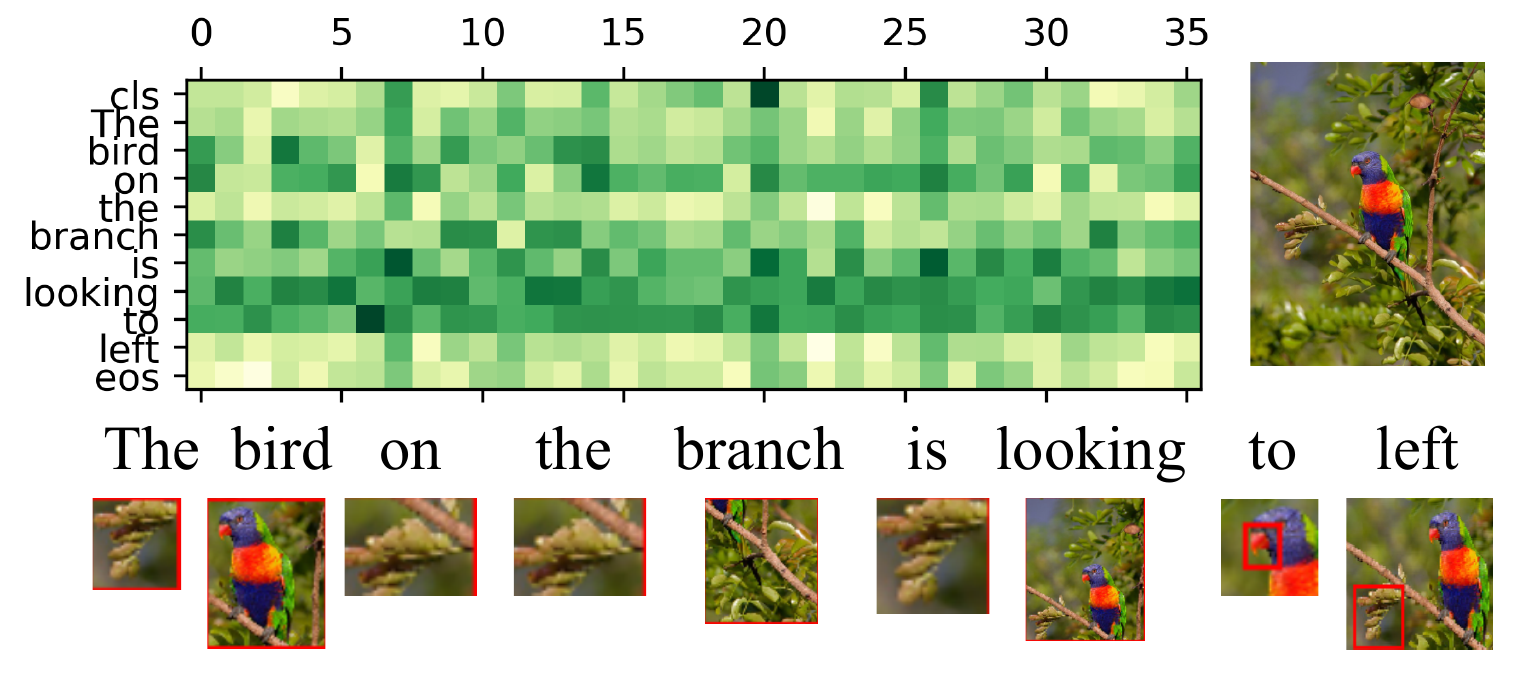}
\caption{The entity affinity matrix between textual (rows) and visual (columns) modalities. The darker color indicates the higher relevance score. The ROIs with maximum relevance score for each word are shown paired with the words.}
\label{fig:entity_matrix}
\end{figure}

\paragraph{Effect of Entity Relevance.}
We remove the entity relevance representation $\Phi_{\mathcal{D}_\mu,\mathcal{D}_\nu}$ from the final feature $\Phi$. As shown in Table~\ref{tab:ablation}, the test accuracy is reduced by $5.4\%$. This is a significant difference of performance among Transformer based models~\cite{li2019visualbert,lu2019vilbert,tan2019lxmert}.
To highlight the significance of entity relevance, we visualize an example affinity matrix in Figure~\ref{fig:entity_matrix}. The two major entities, ``bird'' and ``branch'', are matched perfectly. More interestingly, the three ROIs which are matching the phrase ``looking to left'' capture an indicator (the beak), a direction (left), and the semantic of the whole phrase.

\begin{figure}
\centering
\includegraphics[width=1.0\columnwidth]{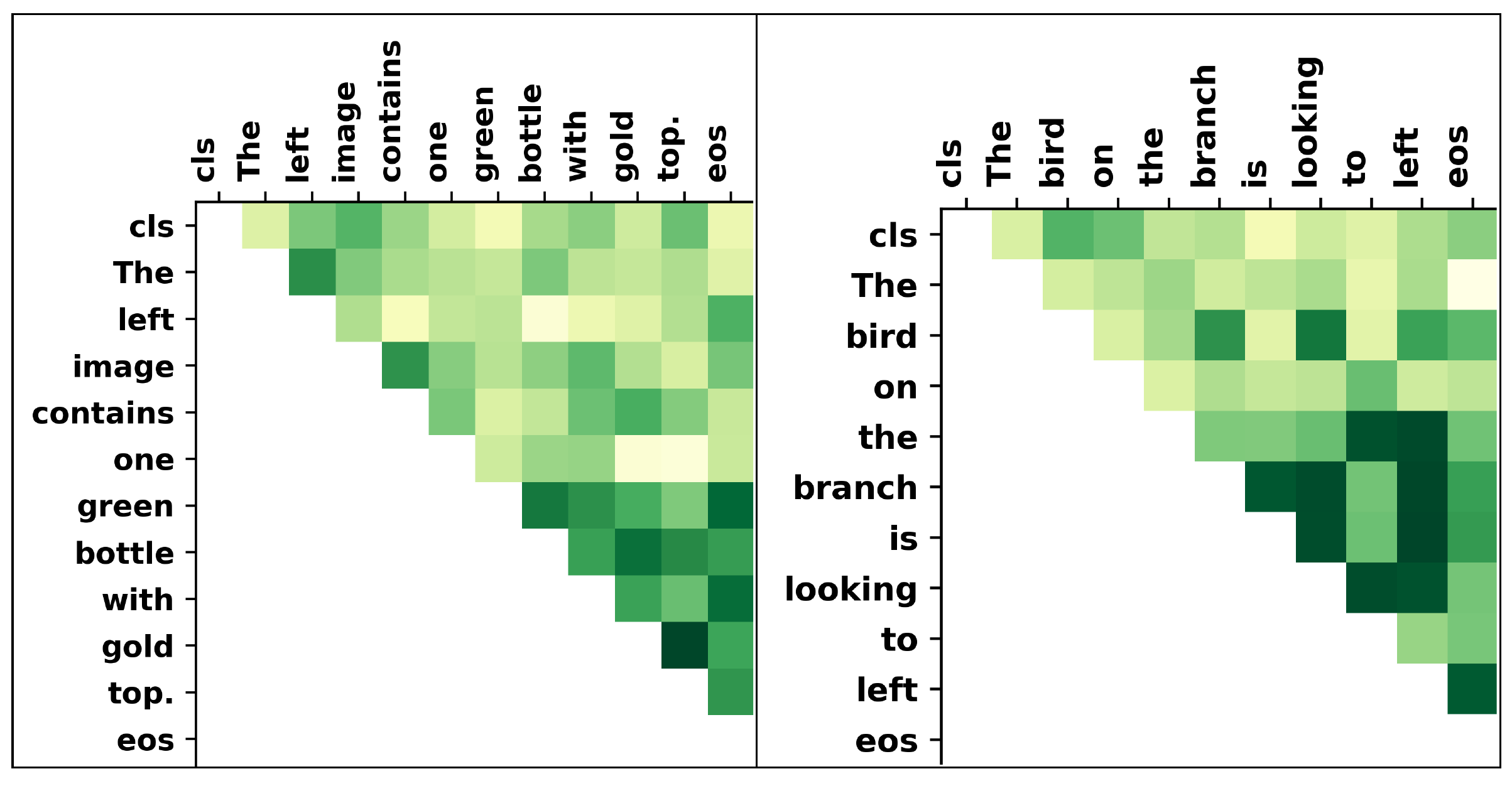}
\caption{The relation ranking score of two example sentence. The darker color indicates the higher ranking score.}
\label{fig:relevance_matrix}
\end{figure}

\paragraph{Effect of Relational Relevance.}
We remove the entity relevance representation $\Phi_{\mathcal{K}_\mu,\mathcal{K}_\nu}$ from the final feature $\Phi$. A $2.5\%$ decrease in test accuracy is observed in Table~\ref{tab:ablation}. We argue that CMR models high-order relations, which are not captured in entity relevance, by modeling relational relevance. We present two examples of textual relation ranking scores in Figure~\ref{fig:relevance_matrix}. The learned ranking score highlights the important pairs, for example ``gold - top'', ``looking - left'', which describe the important relations in textual modality.

\section{Conclusion}

In this paper, we propose a novel cross-modality relevance (CMR) for language and vision reasoning. Particularly, we argue for the significance of relevance between the components of the two modalities for reasoning, which includes entity relevance and relational relevance. We propose an end-to-end cross-modality relevance framework that is tailored for language and vision reasoning. We evaluate the proposed CMR on NLVR and VQA tasks. Our approach exceeds the state-of-the-art on $\mbox{NLVR}^2$  and VQA v2.0 datasets. Moreover, the model trained on $\mbox{NLVR}^2$ boosts the training of VQA v2.0 dataset. The experiments and the empirical analysis demonstrate CMR's capability of modeling relational relevance for reasoning and consequently its better generalizability to unobserved data.
This result indicates the significance of relevance patterns. Our proposed architectural component for capturing relevance patterns can be used independently from the full CMR architecture and is potentially applicable for other multi-modal tasks.

\section*{Acknowledgments}
We thank the anonymous reviewers for their helpful comments.  
This project is supported by National Science Foundation (NSF) CAREER award $\#1845771$.
\bibliography{acl2020}
\bibliographystyle{acl_natbib}

\end{document}